\documentclass[runningheads]{llncs}

\usepackage{eccv}

\usepackage{eccvabbrv}
\usepackage{amsmath}
\usepackage{bm}
\usepackage{makecell}
\usepackage{graphicx}
\usepackage{booktabs}
\usepackage{wrapfig}
\usepackage[accsupp]{axessibility}  
\usepackage[breaklinks,colorlinks,citecolor=eccvblue]{hyperref}
\usepackage{orcidlink}
\usepackage{tikz}

\usepackage{float}
\usepackage{multirow}
\usepackage{pifont}       
\usepackage{bbding}       
\usepackage{xcolor}

\usepackage{amssymb}    

\usepackage{fontawesome}  
 
\newcommand{\cmark}{\ding{51}}

\begin{document}

\title{ExpoMotion: A Large-Scale Benchmark and A Householder Projection Network for Multi-Exposure Fusion}

\titlerunning{ExpoMotion}
\author{Yao Liu\inst{1,3,5}\thanks{Both authors contributed equally to this research.} \and
Lishen Qu\inst{1,3,5}\protect\footnotemark[1] \and
Shihao Zhou\inst{3} \and
Jie Liang\inst{5} \and
Hui Zeng\inst{5} \and
Yabin Peng\inst{3} \and
Huipeng Lin\inst{3} \and
Lei Zhang\inst{4} \and
Jufeng Yang\inst{1,2,3}\thanks{Corresponding author.}
}

\authorrunning{Y.~Liu et al.}

\institute{Nankai International Advanced Research Institute (SHENZHEN·FUTIAN) \and
Pengcheng Laboratory \and
College of Computer Science, Nankai University \and
The Hong Kong Polytechnic University \and
OPPO Research Institute \\
{\tt\small \ \ liuyao@mail.nankai.edu.cn, \ yangjufeng@nankai.edu.cn} \ \\
}
\maketitle


\vspace{-0.2in}
\begin{figure}[h!]
  \centering
  \includegraphics[width=0.95\linewidth]{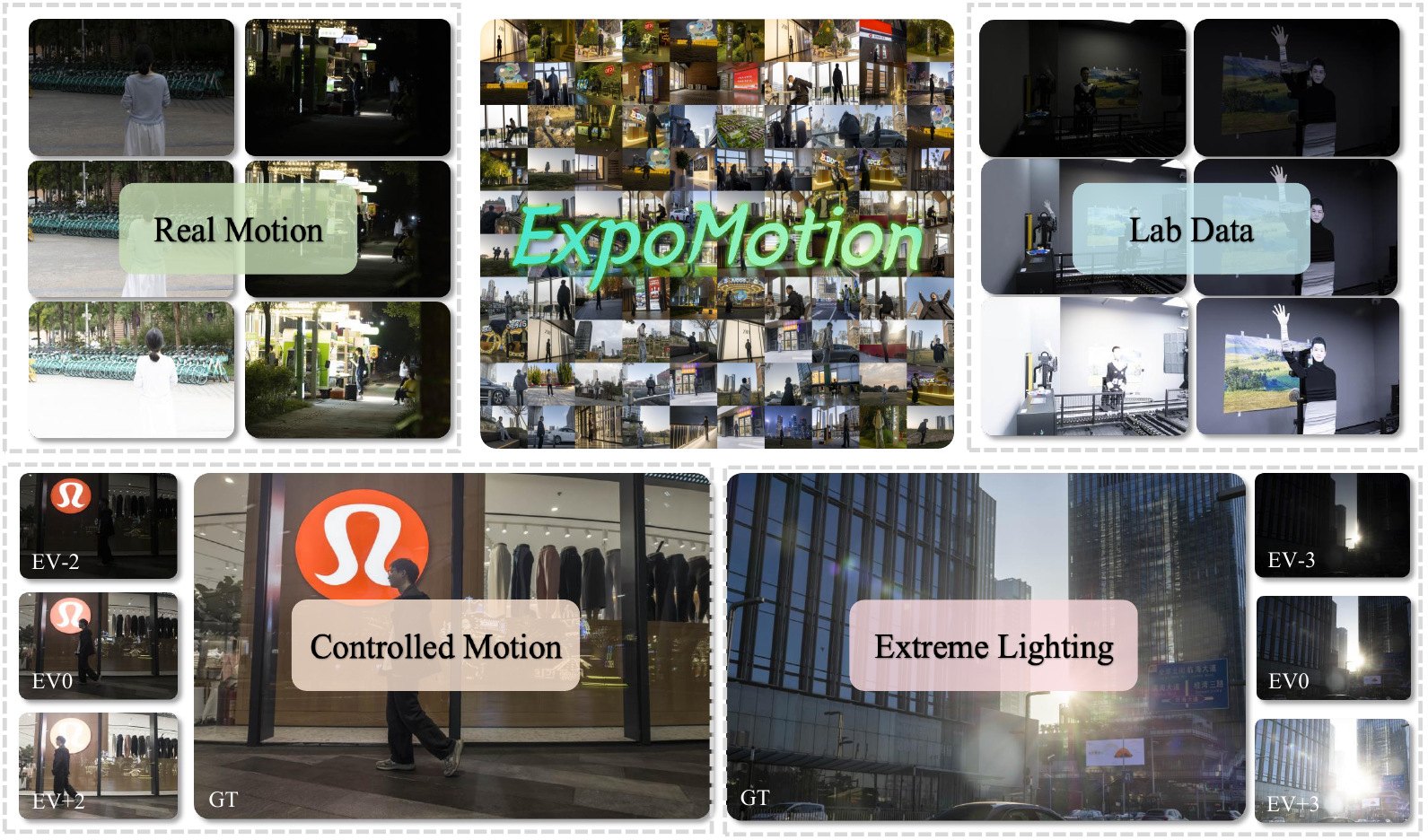}
  \caption{
    Overview of the proposed large-scale MEF dataset. 
    Deviating from the conventional focus on perceptual quality in static scenes, our work introduces a large-scale MEF dataset dedicated to simultaneous detail restoration and motion artifact suppression. We integrate data from controlled and real-world motion, extreme lighting environments, and laboratory setups. 
  }

  \vspace{-0.4in}
  \label{fig:pull}
\end{figure}

\begin{abstract}

Multi-Exposure Fusion (MEF) effectively extends dynamic range, but practical deployment is hindered by motion-induced ghosting and the scarcity of high-quality dynamic benchmarks.
Current benchmarks largely neglect dynamic scenes and lack reliable ground truth, making it difficult to handle the complexity of real-world motions.
In response, we introduce ExpoMotion, a large-scale benchmark designed to evaluate deghosting capabilities.
Comprising 1,738 sequences and 10,909 images across diverse environments, it covers a wide range of motions and provides high-fidelity GTs constructed through an expert-guided acquisition pipeline.
To tackle the complex dynamics and extreme conditions captured in this benchmark, we propose the Householder Orthogonal Projection network (HOP), which revisits MEF deghosting from a mathematical perspective via Householder transformation, decoupling multi-frame alignment into exposure pre-alignment and ghost filtering.
Specifically, the Global Priors Illumination Alignment (GPIA) module first rectifies drastic dynamic range discrepancies by utilizing global statistics for exposure harmonization.
Regarding ghost removal, our Householder Orthogonal Attention (HOA) models artifacts as orthogonal perturbations. By employing a dynamic Householder reflector, HOA effectively projects ghosts out of the feature manifold while preserving high-frequency details.
Experiments demonstrate that our ExpoMotion dataset enables superior generalization and artifact-free detail restoration, while also validating the effectiveness and efficiency of the HOP method.
The dataset and code are available at \url{https://github.com/Leo-LiuYao/ExpoMotion}.

\end{abstract}

\section{Introduction}
\label{sec:intro}

Although real-world scenes exhibit vast dynamic ranges spanning over 10 orders of magnitude, ranging from starlight to direct sunlight, standard imaging sensors possess a limited full-well capacity~\cite{liu2023joint, Bai_2025_ICCV, LearningHDR}. 
This physical limitation imposes a trade-off where preserving highlights crushes shadows, and exposing for shadows blows out highlights. Consequently, computational solutions are required to bridge this perceptual gap~\cite{Debevec1997Recovering,hu2013hdr}.
Two primary paradigms exist to recover these lost details.
HDR Reconstruction recovers a linear radiance map, but requires complex Tone Mapping Operators for display, often introducing secondary artifacts like halos or contrast degradation~\cite{reinhard2023photographic}.
Multi-Exposure Fusion (MEF) offers a compelling alternative by directly synthesizing a high-quality image in the display domain~\cite{mertens2007exposure,grosch2006fast, UltraFusion}.
By bypassing explicit radiance estimation, MEF avoids tone-mapping distortions, aiming to seamlessly blend the most informative features into a perceptually pleasing result suitable for standard displays.

Despite its popularity, MEF struggles significantly in dynamic environments.
Camera shake and moving objects introduce severe misalignments, manifesting as ghosting or tearing artifacts~\cite{wu2018deep,chen2022attention, yan2019attention}. Current benchmarks for multi-exposure fusion primarily focus on static scenes and detail restoration, with few considering performance in dynamic settings. To date, a large-scale, dedicated dataset capable of comprehensively evaluating both deghosting capability and perceptual visual quality in real-world dynamic scenes remains absent~\cite{fang2019perceptual, Wu_2018_ECCV}.
Consequently, a common workaround is to tone-map HDR reconstruction datasets and use the resulting LDR images as supervision for MEF.
While convenient, this practice inevitably inherits the limitations of tone mapping, such as contrast distortion.
%
As a result, the generated supervision often struggles to faithfully reflect perceptual preferences~\cite{zhang2023revisiting, liu2023emef, jiang2023meflut}.

Current benchmarks encounter a dilemma where realistic motion lacks reliable GT, and synthetic motion lacks realism~\cite{shu2024towards, tel2023alignment, Kong_2024_ECCV, cai2018learning, zhang2021benchmarking}.
To address these issues, we introduce ExpoMotion, a large-scale benchmark specifically designed for the joint evaluation of deghosting performance and perceptual visual quality. 
We integrate several established capture strategies to curate a challenging benchmark that features in-the-wild object motion alongside extreme exposure disparities~\cite{heo2010ghost}.
Our dataset encompasses a diverse range of real-world scenarios captured across both daytime and nighttime conditions, including restaurants, beaches, staircases, streets, and other daily-life environments. 
The dataset includes motion types such as controlled motion, for quantitative evaluation, and real-world motion, for practical validity.
For label generation, we first produce candidate ground truths via multiple algorithms, followed by selection and refinement by professional imaging experts. 
This annotation strategy ensures that our ground truths are not only robust to extreme motion artifacts and exposure variations but also align with human perceptual preferences.

Most existing MEF and HDR methods depend on deformable alignment, such as Optical Flow, and standard Attention to aggregate features~\cite{liu2022ghost, liu2023joint, Tel_2023_ICCV}.
However, these approaches face intrinsic limitations. Flow-based warping inevitably introduces geometric distortions in occluded regions~\cite{chan2021understanding, 8989959}. 
More critically, standard attention mechanisms calculate aggregation weights based on feature similarity~\cite{yan2019attention, chen2022attention}.
While this design principle effectively aligns features in general restoration, it faces an intrinsic paradox in MEF. The core objective here is to retrieve complementary details from auxiliary frames, such as textures in over-exposed shadows or under-exposed highlights, that are precisely missing or severely degraded in the reference view~\cite{luo2023multi,zeng2025vision}.
To address this, we augment standard attention with the Householder transformation. By explicitly modeling feature rejection through geometric projection, our method enables the network to distinguish between useful complementary details and misalignment artifacts~\cite{qin2025no,qin2025boosting}.
We reconceptualize MEF alignment not as a single-step feature-matching task but as a decoupled process consisting of exposure pre-alignment and ghost filtering.
This perspective addresses the twin challenges of MEF: the extreme brightness disparities that confuse standard matching metrics, and the motion artifacts that require strict geometric exclusion.
To implement this, we propose the Householder Orthogonal Projection network (HOP).
To address exposure discrepancies, the Global Priors Illumination Alignment (GPIA) module utilizes global statistics to rectify dynamic range variations, thereby pre-aligning auxiliary features to the reference exposure.
Building on this exposure pre-alignment, we introduce Householder Orthogonal Attention (HOA) for deghosting.
%
HOA constructs dynamic reflectors from learned misalignment vectors.
Drawing an analogy to optical mirroring, this facilitates an operation that geometrically reflects auxiliary features to match the reference structure, effectively orthogonally projecting out motion-induced artifacts~\cite{wang2025otlrm,wang2025deep}.
By enforcing this mathematically grounded constraint, HOP effectively improves deghosting fusion.
Our primary contributions are threefold:
\begin{itemize}
    \vspace{-0.20cm}
    \item[$\bullet$] We construct \textbf{ExpoMotion}, a large-scale dynamic MEF benchmark with 1,738 exposure stacks, curated via diverse acquisition strategies across both natural and laboratory settings for rigorous artifact evaluation.
    \item[$\bullet$] We propose the \textbf{Householder Orthogonal Projection (HOP)} network, which decouples alignment by utilizing global priors to harmonize illumination discrepancies and Householder transformations to project out ghosts.
    \item[$\bullet$] Extensive experiments demonstrate that HOP outperforms state-of-the-art methods in both quantitative metrics and perceptual quality.
\end{itemize}

\section{Related Work}
\label{sec:relatedwork}
\noindent \textbf{Multi-Exposure HDR Imaging.}
Existing approaches for high-dynamic-range imaging generally follow two paradigms: HDR reconstruction in the linear domain and Multi-Exposure Fusion (MEF) in the LDR domain~\cite{UltraFusion, qu2026againflexibleframetransformermultiexposure}.
HDR reconstruction methods aim to recover linear irradiance maps by inverting the Camera Response Function (CRF)~\cite{LearningHDR}.
Pioneering deep learning works utilized Convolutional Neural Networks (CNNs) to learn exposure fusion weights directly from data~\cite{LearningHDR, wu2018deep}, establishing a strong baseline.
Later advancements introduced optical flow to explicitly mitigate motion misalignment during the CNN-based merging process~\cite{Kong_2024_ECCV, prabhakar2020towards, Li_2025_ICCV}.
To address the limited receptive field of CNNs, recent research has shifted towards Transformer-based architectures~\cite{chen2022attention, yan2019attention, Tel_2023_ICCV}, employing self-attention mechanisms to model long-range ghosting dependencies~\cite{liu2022ghost, song2022selective}.
Furthermore, generative diffusion models have been explored to hallucinate missing details in saturated regions, treating HDR reconstruction as a conditional generation task~\cite{yan2023toward}.
Alternatively, Multi-Exposure Fusion (MEF) bypasses CRF calibration by directly fusing the LDR sequence to produce a visually pleasing result~\cite{jiang2023meflut, UltraFusion}.
Traditional optimization-based methods focused on crafting weight maps based on hand-crafted features like local contrast and saturation~\cite{mertens2007exposure, li2014selectively, ma2017robust}.
In the deep learning era, MEF has evolved from simple pixel-wise fusion~\cite{ram2017deepfuse, prabhakar2019fast} to sophisticated end-to-end architectures that jointly optimize for structural fidelity and perceptual aesthetics~\cite{xu2020u2fusion, zhao2024image, Bai_2025_ICCV}.
Across both paradigms, the central challenge lies in deghosting—mitigating artifacts caused by camera shake or dynamic objects.
While explicit optical flow~\cite{LearningHDR, wu2018deep} and implicit attention mechanisms~\cite{chen2022attention, yan2019attention} attempt to align features, large displacements and occlusions often break correspondence assumptions, leaving residual artifacts in the fused output.

\noindent \textbf{HDR Datasets and Benchmarks.}
Curating high-quality HDR datasets is notoriously difficult due to the requirement of capturing multiple exposures in rapid succession.
In dynamic environments, unavoidable motion between frames precludes the existence of a perfect, naturally aligned ground truth.
To synthesize labels, earlier works adopted controlled protocols, such as recombining static brackets to simulate stop-motion dynamics~\cite{hu2013hdr, Shu_2024_CVPR}.
A prominent baseline, Kalantari et al.~\cite{LearningHDR}, introduced a dataset with controllable motion, though these synthesized movements often lack the complex motion blur and occlusions found in real-world captures.
Recent efforts utilize graphics engines to generate large-scale synthetic datasets with perfect pixel-wise alignment~\cite{barua2025gta, wang2025s2r}.
However, synthetic data suffers from a distinct domain gap—simulated noise and rendering artifacts often differ significantly from the physics of real sensor optics, limiting generalization.
Furthermore, benchmarks specifically designed for MEF are largely restricted to static scenes~\cite{cai2018learning, zhang2021benchmarking}.
High-quality dynamic datasets that support rigorous evaluation of both HDR reconstruction and MEF deghosting remain scarce~\cite{jiang2023meflut}, underscoring the necessity of more robust evaluation protocols.

\section{Proposed Dataset}

\noindent \textbf{Multi-Exposure Image Collection.}
Existing dynamic HDR datasets often rely on static bracketing to circumvent the difficulty of synthesizing ground truth (GT) from misaligned exposures.
Specifically, Kong \etal~\cite{Kong_2024_ECCV} introduce camera perspective shifts between static captures (Multi-View Static-Bracketing), while Kalantari \etal~\cite{LearningHDR} require subjects to pose in distinct stationary states (Multi-Pose Static-Bracketing).
We adopt the two aforementioned protocols and introduce a third scheme: Static–Dynamic Exposure Bracketing.
We capture two bracketed sets for each scene under identical settings: a static set for synthesizing reliable ground truth, and a dynamic set for input, where the reference frame is swapped with its static counterpart.
This design introduces more motion patterns while maintaining high-quality supervision.
Since these three protocols enforce static scenes during the exposure sequence to ensure artifact-free GT generation, they inherently fail to capture the complexity of diverse, real-world motion.
We refer to the data acquired under these constrained protocols as controlled data.

In addition to the in-the-wild sequences, we captured a complementary subset of data in a professional motion laboratory.
This controlled environment allows us to precisely manipulate motion patterns and illumination conditions, and to record sequences with repeatable scene dynamics. 
Additionally, we captured real-world dynamic scenes, which serve as a benchmark for no-reference perceptual evaluation.
By integrating these strategies, we construct a large-scale MEF dataset comprising 1,738 sequences covering 3, 5, and 7-frame settings with exposures ranging from $\mathrm{EV}{-}3$ to $\mathrm{EV}{+}3$, amounting to a total of 10,909 images.

\noindent\textbf{Data Filtering.}
During data collection, we captured 14{,}235 multi-exposure sequences under different exposure settings. 
Since many sequences contain distorted or low-quality frames, we perform a strict filtering process to curate a high-quality dataset.
We first discard sequences with evident degradations, such as defocus blur, strong sensor noise, noticeable handshake, or other visible imaging artifacts.
Moreover, to facilitate reliable ground-truth (GT) construction, we remove sequences that are not strictly static within each exposure bracket, \emph{i.e.}, any intra-bracket motion that breaks the static-scene assumption required for synthesizing high-quality MEF supervision. 
For repeated sequences, we manually select the one with the best alignment quality and clearest content.
After this stage, more than 80\% of the collected sequences are filtered out, leaving about 2{,}850 candidate sequences.
We then apply representative state-of-the-art MEF methods to these candidates to generate fused reference images, and further prune sequences according to the quality and consistency of the resulting references.

\begin{figure*}[t]
	\begin{center}
		\includegraphics[width=1.0\linewidth]{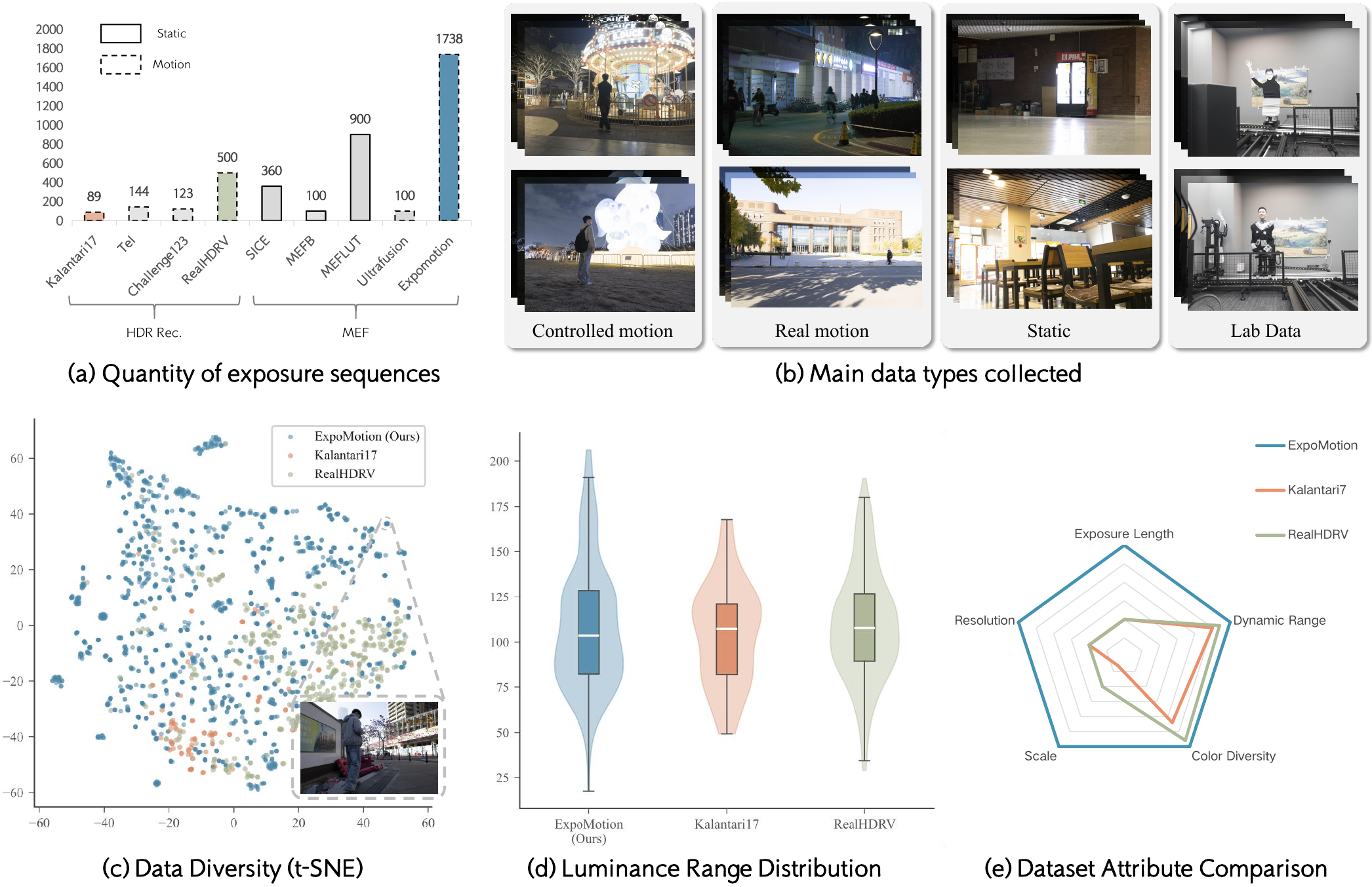}
	\end{center}
 \vspace{-0.4cm}
	\caption{Statistics and diversity of the collected ExpoMotion dataset.
(a) Quantitative comparison showing our dataset is the largest among existing real-world MEF and HDR benchmarks.
(b) Representative samples of the four data acquisition types.
(c) T-SNE visualization showing superior feature diversity compared to related datasets.
(d) Luminance range distribution.
(e) A radar chart highlighting our comprehensive advantages in scale, dynamic range, and color diversity.}
	\label{fig:stats}
    \vspace{-0.4cm}
\end{figure*}

\noindent\textbf{Ground Truth Generation.} 
We construct the GT labels using a human-in-the-loop pipeline that combines automated generation with professional refinement. 
(1) Candidate Generation: 
For every sequence, we generate a diverse set of fusion results using both classic algorithmic baselines~\cite{mertens2009exposure, sen2012robust, ma2019deep,kou2017multi,ma2019deep,li2013image,bruce2014expoblend,li2020fast} and commercial high-dynamic-range software, including Adobe Lightroom and Photomatix.
This ensures a wide search space covering different de-ghosting strategies and tone-mapping operators.
(2) Subjective Selection:
A panel of raters screens these candidates to identify the best initialization, prioritizing results free from motion artifacts and saturation clipping.
(3) Expert Refinement:
The selected optimal priors are then meticulously refined by five professional photographers/imaging experts. 
They manually adjust the tone curves and color balance to recover shadow details and suppress highlight blowout.
This manual intervention is critical for producing GTs that are not only artifact-free but also photorealistic, avoiding the unnatural halos or color shifts often introduced by fully automated pipelines.

\noindent\textbf{Dataset Analysis and Comparison.}
To demonstrate the superiority of our ExpoMotion dataset, we conduct a comprehensive statistical analysis comparing it with existing mainstream HDR and MEF datasets, including Kalantari17~\cite{LearningHDR}, Tel~\cite{tel2023alignment}, 
Challenge123~\cite{Kong_2024_ECCV}, 
RealHDRV~\cite{shu2024towards}, SICE~\cite{cai2018learning}, MEFB~\cite{zhang2021benchmarking}, MEFLUT~\cite{jiang2023meflut} and UltraFusion~\cite{UltraFusion}. ~\cref{fig:stats} provides a holistic visualization of these comparisons.

As illustrated in ~\cref{fig:stats} (a), our dataset contains a total of 1,738 multi-exposure sequences, significantly surpassing existing benchmarks in scale (e.g., $19\times$ larger than Kalantari17 and $3\times$ larger than RealHDRV). 
Unlike previous works that rely heavily on static scenes, 
ExpoMotion features a motion-centric composition that focuses primarily on dynamic sequences (spanning controlled, real-world, and laboratory motion), while retaining a smaller subset of static data.
In addition to capturing diverse in-the-wild real motions, we established a professional laboratory setting to record precise controlled motion and scene dynamics.
To evaluate the semantic richness of the collected scenes, we visualize their distribution using t-SNE. 
~\cref{fig:stats} (c) demonstrates that our dataset (represented by blue points) covers a much broader semantic space compared to Kalantari17 (orange) and RealHDRV (green), indicating superior diversity in scene content and texture.
This wide distribution suggests that models trained on ExpoMotion are likely to achieve better generalization across a broader range of real-world scenarios.
Beyond semantic content, we analyze the photometric properties essential for high-dynamic-range imaging.
~\cref{fig:stats} (d) plots the distribution of luminance ranges. Our dataset exhibits a more extensive distribution with longer tails, indicating that we capture scenes with more extreme lighting conditions (both deep shadows and bright highlights).
Finally, the radar chart in ~\cref{fig:stats} (e) summarizes five key attributes: scale, dynamic range, resolution, exposure length, and color diversity. ExpoMotion fully encompasses the attributes of previous datasets, establishing a new benchmark for dynamic exposure fusion. Specifically, the average image resolution of ExpoMotion is $2563 \times 1896$, which offers approximately $3\times$ the pixel count of Kalantari17 and RealHDRV dataset.

\section{Proposed Method}
\label{sec:method}

Multi-Exposure Fusion (MEF) aims to integrate the high dynamic range information from auxiliary exposures ($\mathbf{I}_{over}, \mathbf{I}_{under}$) into a reference frame ($\mathbf{I}_{ref}$). The core challenge lies in the tension between maximizing information gain and strictly suppressing motion-induced artifacts (ghosting).
In this work, we formalize the MEF task as taking a triplet of differently exposed images as input consisting of an over-exposed frame $\mathbf{I}_{over}$, a reference frame $\mathbf{I}_{ref}$, and an under-exposed frame $\mathbf{I}_{under}$ and generating a single fused image $\mathbf{I}_{pred}$ as output.
The primary goal of this fusion process is to seamlessly amalgamate the distinct details preserved in each exposure: recovering shadow information from $\mathbf{I}_{over}$, highlight details from $\mathbf{I}_{under}$, and retaining the balanced mid-tones from $\mathbf{I}_{ref}$, all while producing a perceptually pleasing result free of motion-induced ghosting artifacts.

\begin{figure*}[t]
	\begin{center}
		\includegraphics[width=1.0\linewidth]{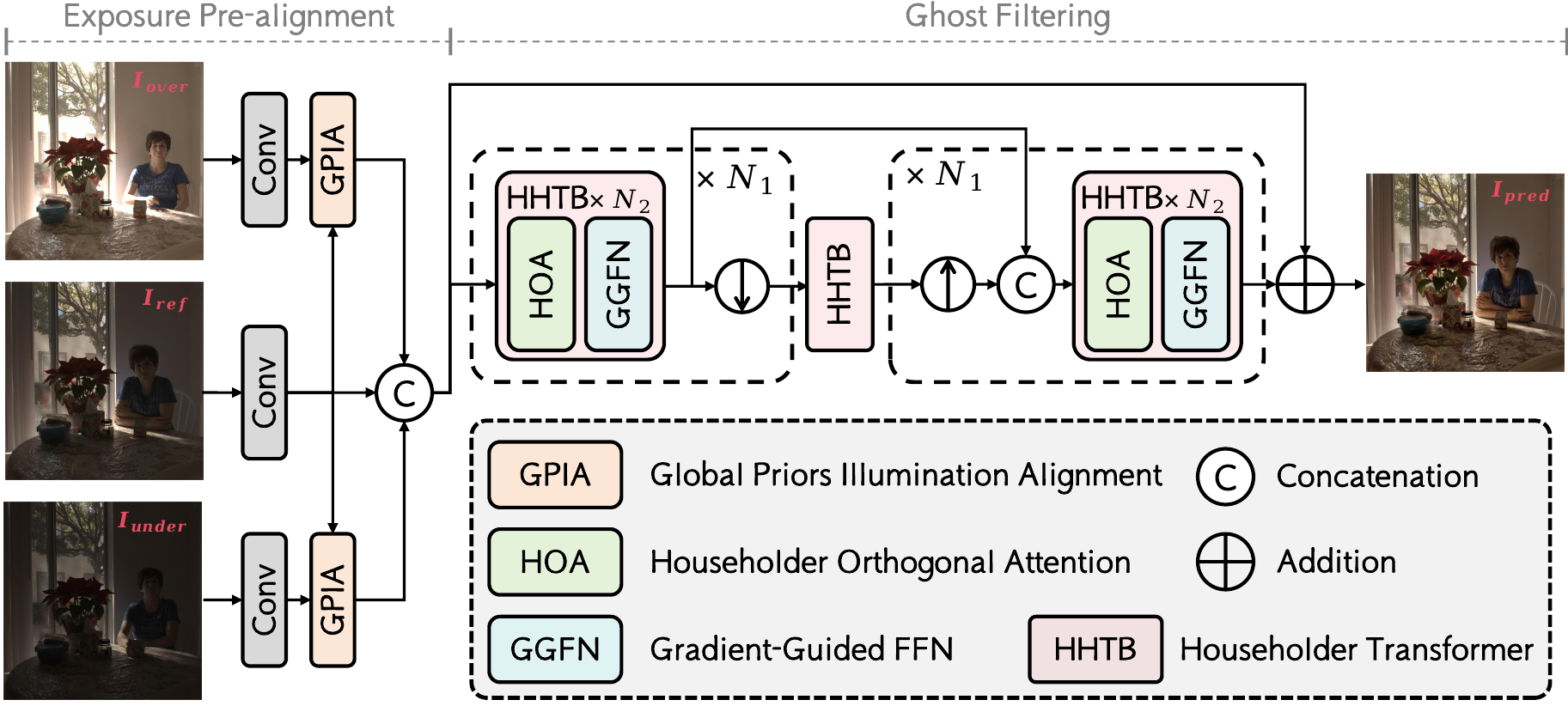}
	\end{center}
    \vspace{-0.4cm}
	\caption{
    Pipeline of the proposed Householder Orthogonal Projection Network (HOP). MEF is decomposed into exposure pre-alignment and ghost filtering stages.
    }
    \vspace{-0.4cm}
	\label{fig:pipeline}
\end{figure*}

To address the issues of inconsistent brightness and motion artifacts across the three input images, we propose the Householder Orthogonal Projection Network (HOP). 
We decompose the MEF task into two core stages: exposure pre-alignment and ghost filtering.
Specifically, we first align the brightness of the two auxiliary frames to the reference frame using global illumination priors. 
Then, inspired by the Householder transformation, we model the artifact removal problem as geometrically mirroring misaligned auxiliary features to match the structural pose of reference features. 
The detailed pipeline is illustrated in ~\cref{fig:pipeline}. We first perform exposure pre-alignment on the three input images. Subsequently, the features of these three images are concatenated and fed into a U-shaped Transformer architecture, where artifact rejection and filtering are conducted within each Transformer block.

\subsection{Global Priors Illumination Alignment (GPIA).}
A major challenge in HDR features is the brightness discrepancy. 
Inspired by the Retinex theory, where illumination is often low-frequency, we introduce the GPIA module to pre-align the features before encoding.
Given the initial feature maps $\mathbf{F}_{Ref}$ and $\mathbf{F}_{Aux}$. We first extract the global illumination priors via average pooling with a $16 \times 16$ kernel:
\begin{equation}
    \mathbf{L}_{Ref} = \phi(\text{AvgPool}(\mathbf{F}_{Ref})), \quad \mathbf{L}_{Aux} = \phi(\text{AvgPool}(\mathbf{F}_{Aux}))
\end{equation}
where $\phi(\cdot)$ is a lightweight mapping function. The illumination ratio map $\mathcal{R}$ is computed to modulate the auxiliary stream:
\begin{equation}
    \mathcal{R} = \frac{\text{Up}(\mathbf{L}_{Ref}) + \epsilon}{\text{Up}(\mathbf{L}_{Aux}) + \epsilon}, \quad \mathbf{F}_{Aux}' = \mathbf{F}_{Aux} \odot \mathcal{R}
\end{equation}

By aligning the exposure levels of the auxiliary frame to the reference beforehand, this module decouples illumination correction from motion handling, allowing subsequent layers to focus exclusively on artifact removal.

\subsection{Householder Orthogonal Attention (HOA)}
\label{sec:hoa}

\begin{wrapfigure}{r}{0.38\textwidth}
    \centering
    \vspace{0pt}
    \includegraphics[width=0.35\textwidth]{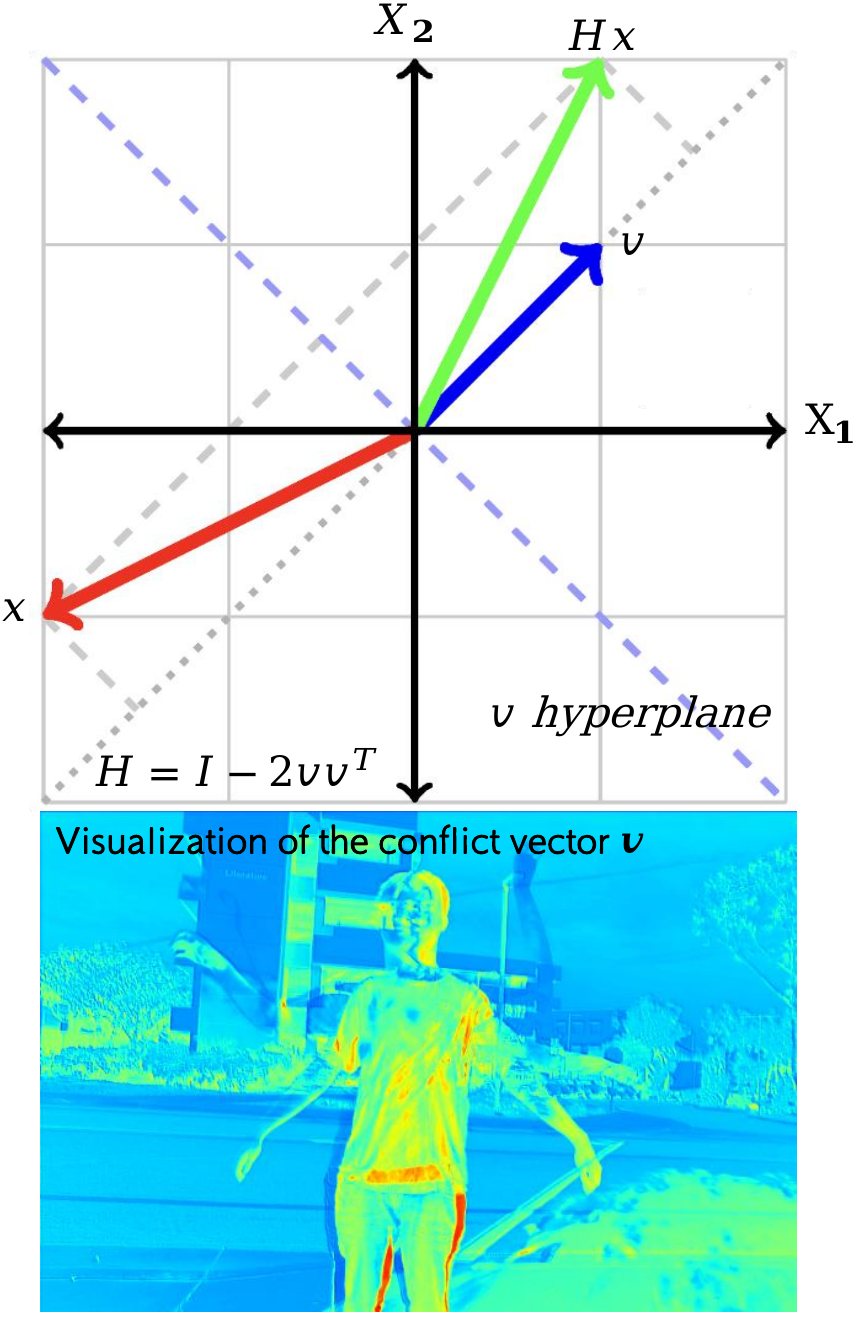}
    \caption{
    Geometric intuition of the Householder transformation.
    }
\label{fig:householder_intuition}
\end{wrapfigure}

The Householder Orthogonal Attention (HOA) module is designed to merge the reference features $\mathbf{F}$ and the pre-aligned features $\mathbf{F}_{SA}$ while discarding misaligned ghosts.
Structurally, it functions as a coarse-to-fine alignment pipeline.
It first employs Transposed Channel Self-Attention to compute the global covariance between frames, which corresponds to $\mathbf{F}_{SA}$ module illustrated in ~\cref{fig:householder_intuition}.
Subsequently, the refined features are passed to our core module: the \textbf{Householder Orthogonal Projection Unit (HOPU)}.
 HOPU treats the deghosting process as a geometric projection problem, utilizing a dynamic Householder reflection matrix to separate the aligned signal subspace from the artifact-inducing noise subspace.

\vspace{0.5em}
\noindent \textbf{Geometric Intuition of Householder Rejection.}
Formally, a Householder matrix $\mathbf{H} \in \mathbb{R}^{d \times d}$ is a fundamental linear algebra operator that performs an orthogonal reflection of a vector across a hyperplane defined by a unit normal vector $\mathbf{v}$. This reflection can be expressed as:
\begin{equation}
    \mathbf{H} = \mathbf{I} - 2\mathbf{v}\mathbf{v}^T
\end{equation}
where $\mathbf{I}$ is the identity matrix.

Inspired by this idea, to adapt this powerful geometric tool for deep feature learning in MEF, we propose an Adaptive Householder-like Projection:
\begin{equation}
\mathbf{H}_{hop} = \mathbf{I} - \alpha \mathbf{v}\mathbf{v}^T
\end{equation}
where $\boldsymbol{\alpha} \in \mathbb{R}^C$ is a learnable scaling vector initialized to zero, allowing the network to fine-tune the rejection intensity per channel. 

Specifically, we define the conflict vector $\mathbf{v}$ based on feature differences. The ghost artifacts in the misaligned auxiliary feature $\mathbf{F}_{SA}$ are theoretically modeled as its projection onto $\mathbf{v}$. By applying $\mathbf{H}_{hop}$, we perform an orthogonal rejection to subtract this component, recovering the clean signal $\mathbf{F}_{clean}$ on the conflict plane. This geometric mirroring operation rigorously filters out motion-induced discrepancies, offering a robust alternative to standard similarity-based aggregation.

\vspace{0.5em}
\noindent \textbf{Difference-Aware Vector Generation.}
Instead of concatenation, we explicitly model the conflict direction $\mathbf{v}$ from the feature difference $\mathbf{D} = \mathbf{F} - \mathbf{F}_{SA}$. This acts as a strong prior for ghost localization:
\begin{equation}
    \mathbf{v}_{raw} = \mathcal{G}_{dw}(\mathbf{D}), \quad \mathbf{v} = \frac{\mathbf{v}_{raw}}{||\mathbf{v}_{raw}||_2 + \epsilon}
\end{equation}
where $\mathcal{G}_{dw}$ denotes a parameter-efficient depth-wise convolution block. Normalization ensures $\mathbf{v}$ lies on the unit hypersphere.

\vspace{0.5em}
\noindent \textbf{Adaptive Orthogonal Filtering.}
To avoid suppressing useful non-ghost details and transfer the linear operator to non-linear, we introduce an intensity gate $\mathcal{M}$ derived from the magnitude of the difference map:
\begin{equation}
    \mathcal{M} = \tanh\left(\text{AvgPool}(|\mathbf{D}|)\right)
\end{equation}
The clean feature $\mathbf{F}_{clean}$ is obtained by projecting $\mathbf{F}_{Aux}$ onto $\mathbf{v}$ and rejecting this component:
\begin{equation}
    \mathbf{F}_{clean} = \mathbf{F}_{Aux} - \alpha \cdot \mathcal{M} \mathbf{v} \odot (\mathbf{v}^\mathsf{T} \mathbf{F}_{Aux})
\end{equation}
This formulation is computationally efficient and removes the misaligned ghost signal defined by $\mathbf{v}$.

\begin{figure*}[t]
	\begin{center}
	\includegraphics[width=1.0\linewidth]{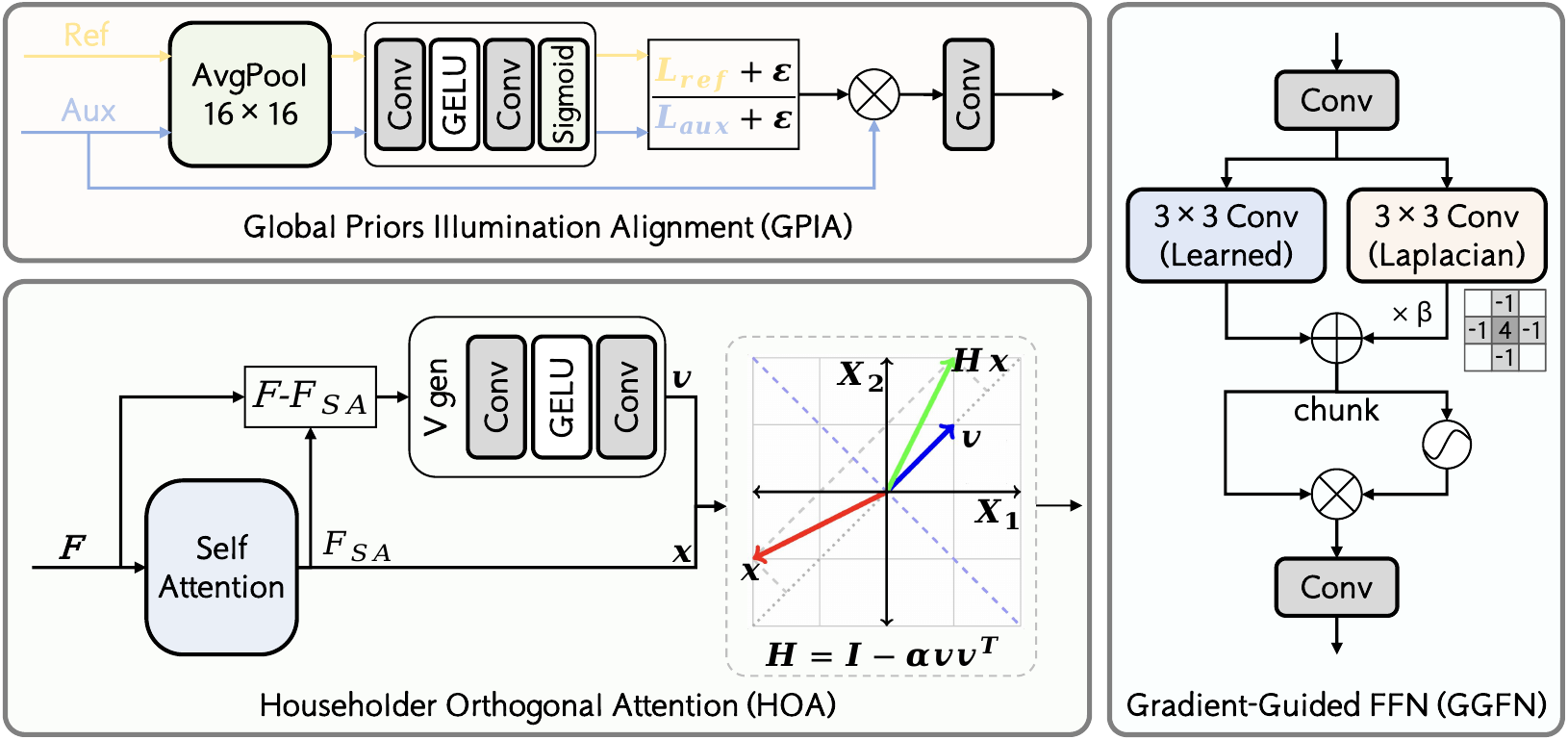}
	\end{center}
    \vspace{-0.4cm}
	\caption{Illustration of HHTB and GPIA. Specifically, HHTB is composed of two core components: Householder Orthogonal Attention (HOA) and Gradient-Guided Feed-Forward Network (GGFN).}
	\label{fig4}
    \vspace{-0.4cm}
\end{figure*}

\subsection{Gradient-Guided Feed-Forward Network (GGFN)}
\label{sec:mtfn}
Standard Feed-Forward Networks (FFNs) typically process features locally but often neglect explicit structural dependencies. However, in HDR deghosting, preserving high-frequency details such as edges and textures is paramount for perceptual quality.
To address this, we propose the Gradient-Guided Feed-Forward Network (GGFN). Instead of relying solely on learned kernels, GGFN synergizes data-driven interpretation with a fixed geometric prior. 

Specifically, the module operates through two parallel pathways:
\begin{itemize}
    \item \textbf{Learned Branch:} A depth-wise convolution ($\text{DWConv}$) captures learnable semantic contexts and local patterns.
    \item \textbf{Laplacian Branch:} 
    We introduce a fixed \textbf{Discrete Laplacian Operator} $\mathcal{L}$ to explicitly extract second-order spatial derivatives (i.e., curvature and boundaries). This branch serves as a hard-coded high-pass filter:
    \begin{equation}
        \mathcal{K}_{Lap} = \begin{bmatrix} 0 & -1 & 0 \\ -1 & 4 & -1 \\ 0 & -1 & 0 \end{bmatrix}
    \end{equation}
\end{itemize}

The Laplacian branch acts as a hard-coded high-pass filter, forcing the network to attend to structural boundaries. The two branches are fused via a learnable mixing parameter $\beta$:
\begin{equation}
    \mathbf{X}_{hidden} = \text{DWConv}(\mathbf{X}_{in}) + \beta (\mathbf{X}_{in} \circledast \mathcal{K}_{Lap})
\end{equation}
Where $\circledast$ denotes cross-correlation, which is implemented as convolution in deep learning libraries. 
By initializing $\beta=0$, the network starts with standard learning and gradually incorporates the Laplacian geometric prior to sharpen textures, effectively regularizing the learning process with explicit edge information.

\section{Experiments}
\subsection{Experimental Setting}

\noindent\textbf{Datasets.}
To evaluate motion-artifact removal and detail restoration for multi-exposure fusion (MEF), we construct two test subsets.
We randomly select 132 samples with ground truth (GT) for full-reference evaluation, and 113 real-world exposure sequences without GT for no-reference evaluation.
The remaining 1,493 exposure sequences are used for training.
Following UltraFusion~\cite{UltraFusion},  we supplement our training data with HDR reconstruction datasets, Kalantari17~\cite{LearningHDR} and RealHDRV~\cite{shu2024towards}.
For both datasets, we apply Photomatix for tone mapping to obtain PNG format fusion targets, and we adopt their official training/testing splits.

\noindent\textbf{Implementation details.}
Our method is implemented in PyTorch and trained end-to-end using the AdamW optimizer. 
%
The initial learning rate is set to $2\times10^{-4}$ and decayed by cosine annealing over 150 epochs.
%
The batch size is set to 4 per GPU.
Automatic mixed precision is enabled.
All experiments are conducted using distributed data-parallel training on 4 NVIDIA V100 GPUs with synchronized batch normalization.
We conduct experiments on two parameter-scale versions of our method: HOP-S ($N_1=2$) and HOP-B ($N_1=3$).

\noindent\textbf{Evaluation metrics.}
For datasets with GT, we report PSNR and SSIM as full-reference metrics.
For datasets without GT, we report MUSIQ~\cite{ke2021musiq}, DeQA-Score~\cite{you2025teaching}, PAQ2PIQ~\cite{ying2020patches}, and HyperIQA~\cite{su2020blindly} as no-reference perceptual metrics.
Specifically, we conduct full-reference evaluations on Kalantari17~\cite{LearningHDR}, RealHDRV~\cite{shu2024towards}, and our controlled-motion testset, 
while performing no-reference evaluations on a composite test set comprising sequences from the Sen~\cite{sen2012robust} and Tursen~\cite{tursun2016objective} datasets, as well as our real-motion testset.

\subsection{Comparison with Previous Work}

\noindent\textbf{Compared methods.}
We compare our approach with representative CNN-based methods, such as SAFNet~\cite{Kong_2024_ECCV} and AFUNet~\cite{Li_2025_ICCV}, and recent Transformer-based restoration architectures, including SCTNet~\cite{tel2023alignment}, Restormer~\cite{zamir2022restormer}, HDR-Trans~\cite{liu2022ghost}, and ASTv2~\cite{zhou2025learning}.
All methods are evaluated under the same protocols when re-training is required.

\begin{table*}[t]\footnotesize
\centering
\caption{Quantitative full-reference results on Kalantari17~\cite{LearningHDR}, RealHDRV~\cite{shu2024towards}, and our controlled-motion test set. Best in \textbf{bold}, second best \underline{underlined}.}
\vspace{-3mm}
\label{tab:fr_results}
\resizebox{\linewidth}{!}{
\begin{tabular}{c|cc|cc|cc|cc}
\toprule
\multirow{2}{*}{Method} &
\multicolumn{2}{c|}{Kalantari17~\cite{LearningHDR}} &
\multicolumn{2}{c|}{RealHDRV~\cite{shu2024towards}} &
\multicolumn{2}{c|}{Expomotion} & 
\multirow{2}{*}{\makecell[c]{Time \\ (ms)}} & 
\multirow{2}{*}{\makecell[c]{Param \\ (M)}} \\
\cline{2-7}
& PSNR$\uparrow$ & SSIM$\uparrow$
& PSNR$\uparrow$ & SSIM$\uparrow$
& PSNR$\uparrow$ & SSIM$\uparrow$\\
\midrule
SAFNet~\cite{Kong_2024_ECCV} (ECCV'24)       & 24.926 & 0.921 & 25.077 & 0.910 & 26.263 & 0.906 & 368 & 1.118 \\
HDR\_Trans~\cite{liu2022ghost} (ECCV'22)& 27.032 & 0.930 & 25.666 & 0.924 & 27.877 & 0.915 & 6355 & 1.454 \\

SCTNet~\cite{tel2023alignment} (ICCV'23)     & 27.100 & 0.931 & 26.420 & 0.923 & 27.806 & 0.912 & 7184 & 3.326  \\

AFUNet~\cite{Li_2025_ICCV} (ICCV'25) & 27.582 & 0.934 & 26.963 & 0.926& 27.672 & 0.917 & 10576 & 1.138  \\

ASTv2~\cite{zhou2025learning} (TPAMI'25)      & 28.113 & 0.934 & 27.563 & 0.937 & 28.573 & 0.925 & 1353 & 7.751 \\

Restormer~\cite{zamir2022restormer} (CVPR'22) & 28.325 & 0.936 & 28.282 & \underline{0.940} & \underline{28.640} & \underline{0.926} & 1860 & 26.10 \\
\midrule
HOP-S (Ours) & \underline{28.501} & \underline{0.940} & \underline{28.285} & 0.937 & 28.593 & \underline{0.926} & 1403 & 3.699  \\

HOP-B (Ours) & \textbf{28.804} & \textbf{0.943} & \textbf{28.704} & \textbf{0.942} & \textbf{28.720} & \textbf{0.927} & 1576 & 15.69  \\
\bottomrule
\end{tabular}}
\end{table*}
\begin{figure*}[t]
	\begin{center}
		\includegraphics[width=1.0\linewidth]{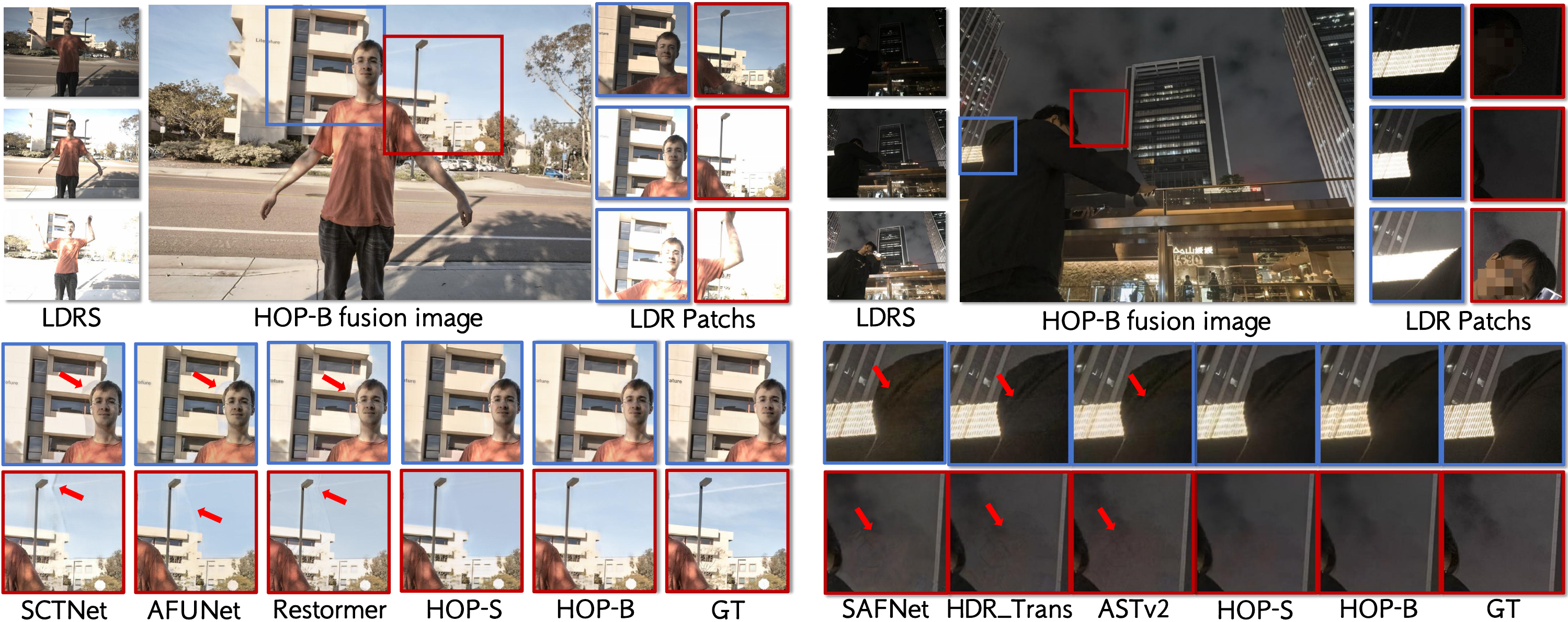}
	\end{center}
    \vspace{-0.3cm}
	\caption{Visual comparisons on Kalantari17~\cite{LearningHDR} dataset and ours. Zoom in for a better view.}
    \vspace{-0.4cm}
	\label{fig:vis}
\end{figure*}

\noindent\textbf{Full-reference comparison.}
\Cref{tab:fr_results} summarizes the quantitative results on the Kalantari17~\cite{LearningHDR}, RealHDRV~\cite{shu2024towards}, and our controlled-motion testsets.
As presented, our HOP-B establishes a new state-of-the-art across all metrics.
Notably, it outperforms the strong competitor Restormer~\cite{zamir2022restormer} by approximately 0.5 dB in PSNR on Kalantari17~\cite{LearningHDR}, while requiring only about 60\% of its Params (15.69M vs. 26.10M).
Even our lightweight variant, HOP-S, achieves remarkable efficiency-performance trade-offs: it surpasses Restormer~\cite{zamir2022restormer} and ASTv2~\cite{zhou2025learning} on the Kalantari17~\cite{LearningHDR} and RealHDRV~\cite{shu2024towards} dataset with significantly reduced computational costs (only 3.699M Params).
Qualitatively, as shown in \cref{fig:vis}, previous methods like AFUNet~\cite{Li_2025_ICCV} and Restormer~\cite{zamir2022restormer} suffer from noticeable ghosting artifacts in dynamic regions.
Specifically, as indicated by the red arrows in the left sub-figure, competing models struggle to distinguish foreground structures from moving backgrounds, resulting in severe ghosting artifacts and unnatural distortions around the lamp pole.
Furthermore, in the challenging night scene on the right, while other models exhibit severe noise and blurring in the dark wall area, our model faithfully recovers clear details in both over-exposed light grids and under-exposed shadows.
In contrast, our method significantly outperforms previous approaches in artifact removal and dark-region denoising.

\begin{table*}[t]\footnotesize
\centering
\caption{Cross-dataset generalization with no-reference metrics. Models are trained on different datasets and tested on (a) Sen~\cite{sen2012robust} and Tursen~\cite{tursun2016objective}, and (b) our real-motion testset. }
\vspace{-3mm}
\label{tab:cross_dataset}
\resizebox{\linewidth}{!}{
\begin{tabular}{c|c|ccccc|ccccc}
\toprule
\multirow{2}{*}{Trainset} & \multirow{2}{*}{Method} &
\multicolumn{5}{c|}{Test on Sen~\cite{sen2012robust} and Tursen~\cite{tursun2016objective}} &
\multicolumn{5}{c}{Test on Ours Real-motion Testset} \\
\cline{3-12}
& &ID& MUSIQ$\uparrow$ & DeQA$\uparrow$ & PAQ2PIQ$\uparrow$ & HyperIQA$\uparrow$
  &ID& MUSIQ$\uparrow$ & DeQA$\uparrow$ & PAQ2PIQ$\uparrow$ & HyperIQA$\uparrow$ \\
\midrule
\multirow{2}{*}{Kalantari17}
& Restormer~\cite{zamir2022restormer} &(a)& 57.45 & 3.214 & 67.98 & 0.4211 &(g)& 59.94 & 3.544 & 70.28 & 0.4314 \\
& HOP-B  &(b)& 57.87 & 3.225 & 68.68 & 0.4328 &(h)& 61.08 & 3.55 & 70.83 & 0.4354                            \\

\midrule
\multirow{2}{*}{RealHDRV}
& Restormer~\cite{zamir2022restormer} &(c)& 60.50 & 3.366 & 69.09 & 0.4736 &(i)& 61.85 & 3.599 & 70.92 & 0.4703 \\
& HOP-B &(d)& 60.89 & 3.395 & 69.97 & 0.4819 &(j)& 62.84 & 3.61 & 71.23 & 0.4824 \\

\midrule
\multirow{2}{*}{ExpoMotion}
& Restormer~\cite{zamir2022restormer} &(e)& 62.32 & 3.535 & 70.96 & 0.4839 &(k)& 65.40 & 3.867 & 72.19 & 0.5195 \\
& HOP-B &(f)& 62.74 & 3.546 & 71.34 & 0.4861 &(l)& 65.68 & 3.87 & 72.19 & 0.5284 \\
\bottomrule
\end{tabular}}
\end{table*}

\begin{figure}[t]
    \centering
    \includegraphics[width=0.95\linewidth]{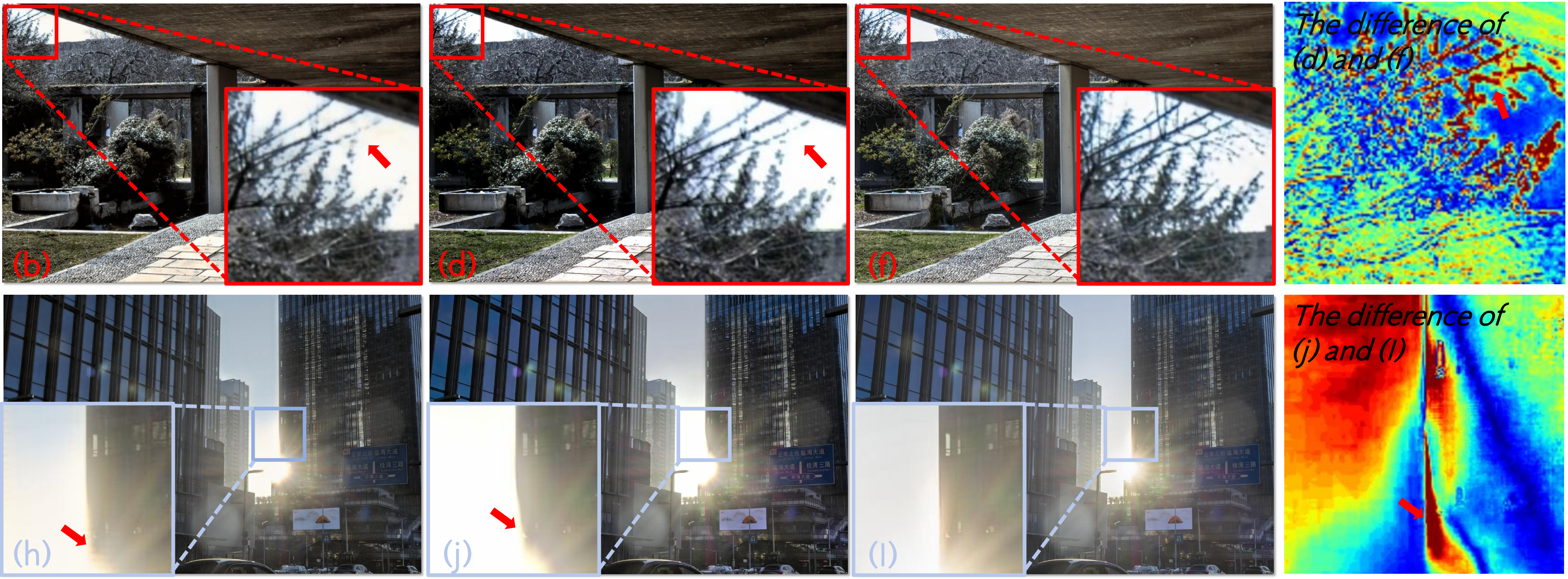}
    \caption{Visual comparison of cross-dataset generalization. The indices of the sub-figures are consistent with those in ~\cref{tab:cross_dataset}.}
    \label{fig:nogt}
    \vspace{-0.3cm}
\end{figure}

\noindent\textbf{No-reference comparison.}
\cref{tab:cross_dataset} presents the perceptual evaluation using non-reference metrics on two real-world test sets (Sen~\cite{sen2012robust}/Tursen~\cite{tursun2016objective} and ours), which lack ground truth.
This experiment effectively validates the generalization capability of models trained on different datasets.
Crucially, the results highlight the superiority of our ExpoMotion dataset.
Specifically, on the Sen and Tursen testset, the HOP-B model trained on our dataset (f) achieves a MUSIQ score of 62.74, outperforming the same model trained on Kalantari17 (b, 57.87) and RealHDRV (d, 60.89) by significant margins of approximately 4.9 and 1.9 points, respectively. 
Moreover, even under identical training data, our HOP-B architecture consistently surpasses Restormer (e.g., comparing e vs. f), demonstrating the effectiveness of our design in rendering visually pleasing HDR images.
~\cref{fig:nogt} presents qualitative comparisons. 
As observed in the top row, competing datasets (~\cref{fig:nogt} (b) and (d)) trained on smaller datasets struggle to handle complex motions, resulting in blurred tree branches and loss of high-frequency details. 
Similarly, in the challenging extreme-lighting scene shown in the bottom row, these methods (~\cref{fig:nogt} (h) and (j)) fail to preserve the geometric integrity, resulting in noticeable structural distortions near strong light sources.
This significant improvement indicates that the diverse motion patterns and realistic luminance variations captured in our dataset enable the network to learn more generalized features effectively.

\subsection{Ablation Study}
\label{sec:ablation}

We utilize the Kalantari17 dataset to dissect the contribution of each component within our lightweight variant, HOP-S. 
As detailed in ~\cref{tab:ablation}, the baseline (row a), devoid of our proposed modules, yields a PSNR of 27.45 dB.

\noindent \textbf{Impact of Individual Modules.} 
Integrating modules individually, like GPIA (b), HOPU (c), or GGFN (d), consistently boosts PSNR by approximately 0.4 dB compared to the baseline. 
Crucially, these gains incur negligible computational overhead ($\leq$0.3 M parameters), verifying the efficiency of our modular design.
As shown in \cref{fig:abvis}, in the top row, (b) with GPIA exhibits better noise recovery on human faces compared to (a). In the bottom row, (g) with HOPU demonstrates more significant artifact removal. These improvements validate the effectiveness of both the GPIA and HOPU modules.

\noindent \textbf{Synergy and Full Model.}
The results further highlight the complementary nature of our components. 
Pairwise combinations (rows e, f) exhibit clear performance jumps over standalone configurations, confirming that the modules effectively collaborate to handle different aspects of restoration.
Consequently, the full model (row g), which integrates GGFN, GPIA, and HOPU, achieves the peak performance of 28.50 dB with a substantial 1.05 dB improvement over the baseline.
Remarkably, this significant gain is achieved with only a marginal 9.3\% increase in Params (from 3.3818 M to 3.6990 M), demonstrating a highly favorable trade-off between reconstruction quality and computational cost.

\begin{table}[t]  
    \setlength{\tabcolsep}{4pt}  
    \footnotesize  
    \begin{minipage}{0.65\linewidth}
        \centering  
        \caption{Ablation study on Kalantari17 dataset. We compare the effects of different combinations of GPIA, HOPU, and GGFN modules, with a total of seven experimental groups from (a) to (g). The GDFN module proposed in Restormer~\cite{zamir2022restormer}.}
        \vspace{-3mm}
        \label{tab:ablation}
        \resizebox{\linewidth}{!}{
            \begin{tabular}{c|cccc|ccc}
                \toprule
                \multirow{2}{*}{ID} & \multirow{2}{*}{GPIA} & \multirow{2}{*}{HOPU} & \multicolumn{2}{c|}{FFN} & \multirow{2}{*}{PSNR} & \multirow{2}{*}{SSIM}  & \multirow{2}{*}{\makecell[c]{Params \\ (M)}} \\
                \cline{4-5}
                &  &  & GDFN & GGFN &  &  &  \\
                \midrule
                (a) & & & \cmark & & 27.4526 & 0.9287  & 3.3818 \\
                (b) & \cmark &  & \cmark & & 27.8562 & 0.9370  & 3.3819 \\
                (c) &  & \cmark &\cmark &  & 27.9003 & 0.9323 &  3.6332 \\
                (d) &  & &  & \cmark & 27.8332 & 0.9376  & 3.4475 \\
                
                
                (e) &  & \cmark&  & \cmark& 28.2321 & 0.9391  & 3.6989 \\
                (f) &\cmark & &  &\cmark & 28.1334 & 0.9382  & 3.4476 \\
                (g) & \cmark & \cmark &  &\cmark & \textbf{28.5011} & \textbf{0.9404}  & 3.6990 \\
                \bottomrule
            \end{tabular}
        }
    \end{minipage}
    \hfill  
    \begin{minipage}{0.3\textwidth}
        \centering
        \includegraphics[width=\textwidth]{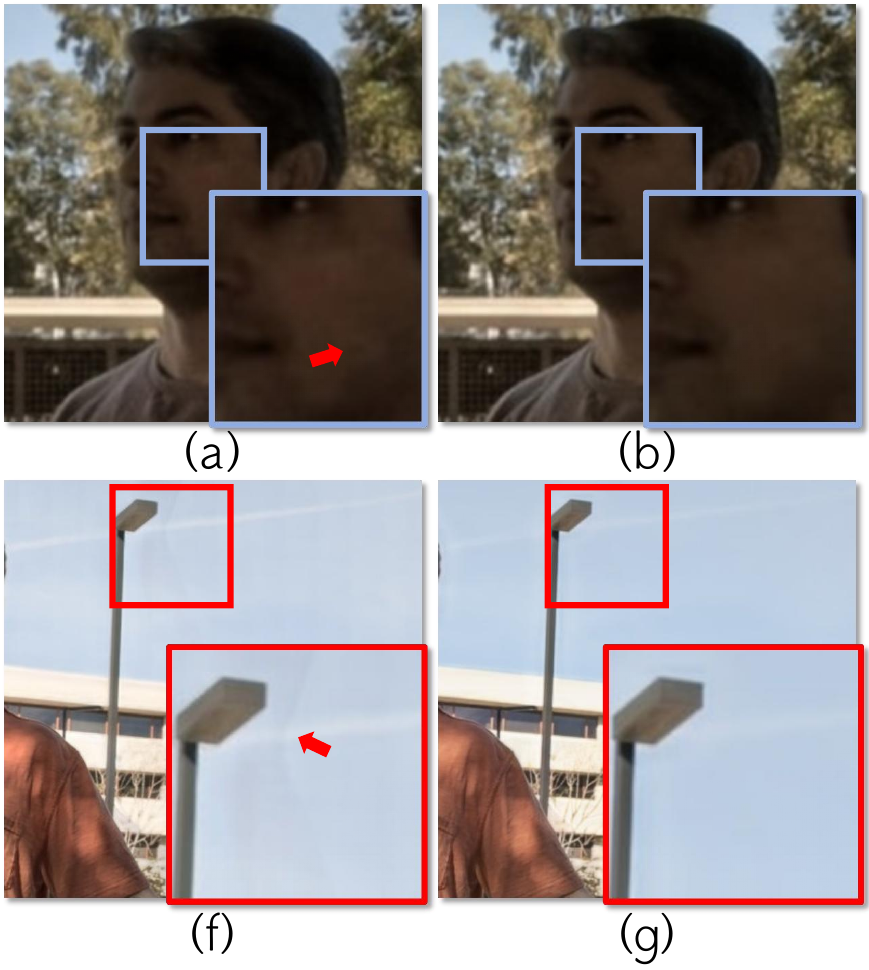}
        \vspace{-3mm}
        \caption{Visual results of the ablation study}
        \label{fig:abvis}
    \end{minipage}
    \vspace{-0.5cm}
\end{table}

\section{Conclusion}
\label{sec:conclusion}

In this work, we address the twin challenges of benchmarking and deghosting in Multi-Exposure Fusion (MEF) for dynamic scenes. 
To overcome the scarcity of reliable evaluation benchmarks, we construct ExpoMotion, a large-scale benchmark that assesses deghosting capabilities, detail restoration, and perceptual aesthetics via a hybrid of controlled and in-the-wild motion sequences. 
On the methodological front, we depart from standard correspondence learning and introduce the Householder Orthogonal Projection (HOP) network.
By mathematically reframing deghosting as a geometric filtering problem, HOP effectively decouples the task: it first harmonizes exposure disparities through global statistics and subsequently projects out motion artifacts using a relaxed Householder transformation.
This coherent design yields superior fusion fidelity and computational efficiency compared to state-of-the-art alternatives.

\noindent \textbf{Limitation.} Our model adopts a three-input fusion framework. This constraint limits the applicability of our method to datasets or devices that provide a flexible number of exposures. Future research will focus on generalizing the network design to accommodate variable input frame counts, further improving the robustness and versatility of the system.

\section{Acknowledgments}
This work was supported by the Shenzhen Science and Technology Program (No. JCYJ20240813114229039), PCL Major Key Project of PCL2025A17-2, the Natural Science Foundation of Tianjin, China (No. 24JCZXJC00040), the National Natural Science Foundation of China (No. 624B2072), the Doctoral Student Program of the Young S\&T Talents Cultivation Project, CAST, the Fundamental Research Funds for the Central Universities (No. 63263253), the Supercomputing Center of Nankai University (NKSC), and OPPO Research Fund.

\bibliographystyle{splncs04}
\bibliography{main}

\end{document}